\documentclass{article}

\usepackage[a4paper, margin=1in]{geometry}
\usepackage{graphicx}  
\usepackage{amsmath}   
\usepackage{amssymb}   
\usepackage{booktabs}  
\usepackage{hyperref}  
\usepackage{cite}    
\usepackage{url}       
\usepackage{float}    
\usepackage{caption}   
\usepackage{subcaption} 
\usepackage{indentfirst}
\usepackage{xcolor}

\title{\textbf{Semantic segmentation for building wooden cube houses}}
\author{Ivan Beleacov\\Independent Researcher}
\date{}

\begin{document}
\sloppy

\maketitle
\begin{abstract}
\bfseries
Automated construction is one of the most promising areas that can improve efficiency, reduce costs and minimize errors in the process of building construction. In this paper, a comparative analysis of three neural network models for semantic segmentation, U-Net(light), LinkNet and PSPNet, is performed. Two specialized datasets with images of houses built from wooden cubes were created for the experiments. The first dataset contains 4 classes (background, foundation, walls, roof) and is designed for basic model evaluation, while the second dataset includes 44 classes where each cube is labeled as a separate object.
The models were trained with the same hyperparameters and their accuracy was evaluated using MeanIoU and F1 Score metrics. According to the results obtained, U-Net(light) showed the best performance with 78\% MeanIoU and 87\% F1 Score on the first dataset and 17\% and 25\% respectively on the second dataset. The poor results on the second dataset are due to the limited amount of data, the complexity of the partitioning and the imbalance of classes, making it difficult to accurately select individual cubes. In addition, overtraining was observed in all experiments, manifested by high accuracy on the training dataset and its significant decrease on the validation dataset.
The present work is the basis for the development of algorithms for automatic generation of staged building plans, which can be further scaled to design complete buildings. Future research is planned to extend the datasets and apply methods to combat overfitting (L1/L2 regularization, Early Stopping). The next stage of work will be the development of algorithms for automatic generation of a step-by-step plan for building houses from cubes using manipulators.

Index Terms-Deep Learning, Computer vision, CNN, Semantic segmentation, Construction materials
\end{abstract}

\section{Introduction}
Automated construction is one of the most promising areas for improving safety, reducing costs and speeding up building construction processes. In addition, automation makes it possible to implement projects in challenging environments, including extreme climate zones, remote regions, and space. However, the effective implementation of automated construction systems requires technologies that can analyze and interpret construction data, including images and drawings.
In recent years, the development of artificial intelligence has greatly enhanced the capabilities of computer vision, making it a key tool for design and automated construction tasks. One of the important challenges remains the development of a system capable of automatically generating a step-by-step plan for the construction of a building, similar to the instructions for assembling a constructor. To solve this problem, deep learning models capable of analyzing images and highlighting the structural elements of a building are applied.
This paper proposes the development of the first stage of a prototype system that uses semantic segmentation techniques to classify the structural elements of a building in order to automatically generate its construction plan.
As an experimental platform, construction from 5x5x5 wooden cubes using manipulators is considered.This approach allows testing the algorithms in a controlled environment, minimizing the computational cost and complexity of data preparation. The results obtained can be scaled for real-world construction applications, where the algorithms will be able to generate instructions tailored to material and architectural requirements.
In the field of computer vision, both classical machine learning algorithms and deep neural networks have traditionally been applied to construction material classification tasks.However, the existing methods solve highly specialized problems and are not oriented to complex automation of design processes. 
At the same time, advances in deep learning have made it possible to apply convolutional neural networks (CNNs) to image segmentation tasks, significantly improving the accuracy of construction object analysis.
In this paper, a comparative analysis of three CNN architectures - U-Net(light), LinkNet and PSPNet - for the task of semantic segmentation of construction cubes based on two specially trained datasets is performed.

\maketitle
\section{Literature Review}
Semantic segmentation \cite{csurka2023semantic} is a computer vision method that not only detects objects in images but also determines their exact spatial location by classifying each pixel. This technique is useful when an image needs to be divided into multiple categories (multiclass classification) rather than just distinguishing between two classes (binary classification).
Convolutional neural networks (CNNs) \cite{oshea2015cnn} were specifically developed for image processing and have shown higher accuracy than traditional multilayer perceptrons (MLPs), while also requiring fewer trainable parameters. A major breakthrough in CNN-based image classification happened with the introduction of AlexNet \cite{krizhevsky2012imagenet}, which won the ImageNet competition in 2012. Later, more advanced models were developed, including VGG \cite{simonyan2014vgg}, ResNet \cite{he2015resnet}, and GoogleNet \cite{szegedy2014googlenet}. However, these models were mostly designed for whole-image classification rather than pixel-level segmentation.
For semantic segmentation, the U-Net architecture \cite{ronneberger2015unet} was introduced. Originally designed for medical image analysis, U-Net has an encoder-decoder structure, where the encoder extracts image features, and the decoder reconstructs spatial information to produce a segmentation map. Other architectures have also been proposed, such as LinkNet \cite{chaurasia2017linknet}, which is optimized for real-time applications, and PSPNet \cite{zhao2017pspnet}, which uses a pyramidal pooling structure (PPM) to capture both local and global image features.
Semantic segmentation has been successfully applied in many fields. For example, U-Net was used for automated urban planning map digitization \cite{guo2018urban}, achieving 99.36\% accuracy and a Jaccard coefficient of 93.63\%. In historical map processing, where annotated data is limited, a technique called age-tracing \cite{yuan2025historical} was used. The method involved training U-Net on a single labeled map and then generating pseudo-labels for similar maps from other time periods, achieving 77.3\% MeanIoU and 97\% accuracy.
Another study \cite{ekim2021roadmaps} improved historical road map segmentation using U-Net++ with a ResNeXt50\_32×4d encoder, resulting in 41.99\% IoU and 46.61\% F1-score. For landslide detection, researchers tested U-Net, LinkNet, PSPNet, and FPN on the Bijie Landslide dataset (770 images with landslides, 2003 without). LinkNet achieved the best results with an F1-score of 85.7\% \cite{oak2024landslide}. This study also showed that CNN-based segmentation methods can significantly reduce processing time and costs compared to traditional approaches.
Beyond segmentation, research has also focused on building material classification. One study \cite{rashidi2015materials} compared SVM, RBF, and MLP for binary classification of construction materials (e.g., concrete, red brick, OSB panels), achieving precision and recall rates between 40\% and 100\%. Another study \cite{mahamia2020materials} used deep learning models such as ResNet152, VGG-16, DenseNet, and NasNet Mobile, where VGG-16 achieved the highest accuracy (above 97\%).
For automatic concrete damage detection, researchers applied Mask R-CNN with transfer learning \cite{kumar2021concrete}, achieving 95.13\% accuracy on a dataset of 800 images and 96.87\% on images from the internet. The authors \cite{park2024semantic} applied semantic segmentation to 3D images for automatic recognition of heavy construction equipment at excavation sites. Four deep learning algorithms (RandLA-Net, KPConv Rigid, KPConv Deformable and SCF-Net) were selected and tested using 3D digital maps to solve the problem. Experimental results showed that the accuracy was lower than similar algorithms for 2D images, but the authors plan to improve the methods in the future. Although this work uses semantic segmentation in construction, the neural networks and tasks are customized specifically to work with 3D images, which may contribute to the development of an automated building plan generation project, but it is not a primary goal at this time.

Although CNNs are widely used for image classification, semantic segmentation is still underutilized in construction-related applications. However, its successful use in other domains and some in construction suggests it could be useful for solving problems requiring precise object segmentation. The U-Net architecture, in particular, is well suited for such applications due to its efficiency on small datasets and its ability to accurately separate objects at the pixel level.
On the other hand, instance segmentation models such as Mask R-CNN and the R-CNN family are capable of distinguishing individual instances within the same class, but they require more computational power.\cite{hafiz2020instanceseg}. Since the objects in this project (wooden cubes) are almost identical in appearance, instance segmentation would not provide significant benefits but would increase complexity. For this reason, semantic segmentation is the best choice, as it effectively captures spatial relationships between different parts of the image while remaining computationally efficient.

\maketitle
\section{Proposed Methods}
\subsection{Dataset}

Own images were used to create the datasets as the data requirements were specific. Two datasets consist of images of 5 different houses built from wooden cubes measuring 5x5x5 cm. The number of cubes in each house varies from 1 to 43. All images were taken with a Google Pixel 6 Pro phone mounted on a tripod.
The shooting of each house was divided into 4 stages: foundation, walls, foundation and walls, and the whole house. Additionally, images were taken of houses with some cubes missing in the construction. This was done to increase the diversity of the data to improve the stability of the neural network training. At each stage, the object was photographed at angles of 0°, 30°, 60°, and 90°. Since houses are symmetrical, only 5 unique angles were photographed, excluding mirror reflections. The number of original images in the first dataset was 374 and 174 in the second dataset. All original images were then labeled using the CVAT.ai tool. The size of the dataset was small and a manual approach was used to accurately mark up each image.
One of the main differences between the two datasets is the number of classes and the partitioning principle. The first dataset of 374 images contains 4 classes: 0 - background, 1 - foundation, 2 - walls, 3 - roof. In the second dataset, it was decided to mark each cube as a separate class, resulting in 44 classes (43 - cubes and 0 - background).This distinction was made intentionally. The first dataset with 4 classes is used to check the quality of images and identify possible problems with markup or snapshots. The second dataset is used for detailed segmentation of each cube. It lays the foundation for an automated house building system where it is important to handle each element separately.
To enable the network to distinguish the roof from the foundation, an additional layer of cubes was added to the roof of the huts, otherwise both elements would look like flat surfaces.
Since the size of the datasets was insufficient for stable training, augmentation was applied using the albumentations library in Python. This increased the number of images in the first dataset to 2149 and the number of images in the second dataset to 870. The following augmentations were used: CLAHE, RandomRotate90, Transpose, ShiftScaleRotate, Blur, OpticalDistortion, GridDistortion and HueSaturationValue. All original photos were taken on a white background and under yellow lighting. Expanding the shooting conditions (variety of backgrounds, lighting) would require an additional increase in the number of original images, which is planned to be done in the future.

\subsection{U-Net(light)}

U-Net was originally introduced in 2015 at the Cell Segmentation Competition. It was initially used for medical image segmentation, but over time it has been used in other applications due to its encoder-decoder architecture (Figure 1).

\begin{figure}[H] 
\centering
\includegraphics[width=0.5\textwidth]{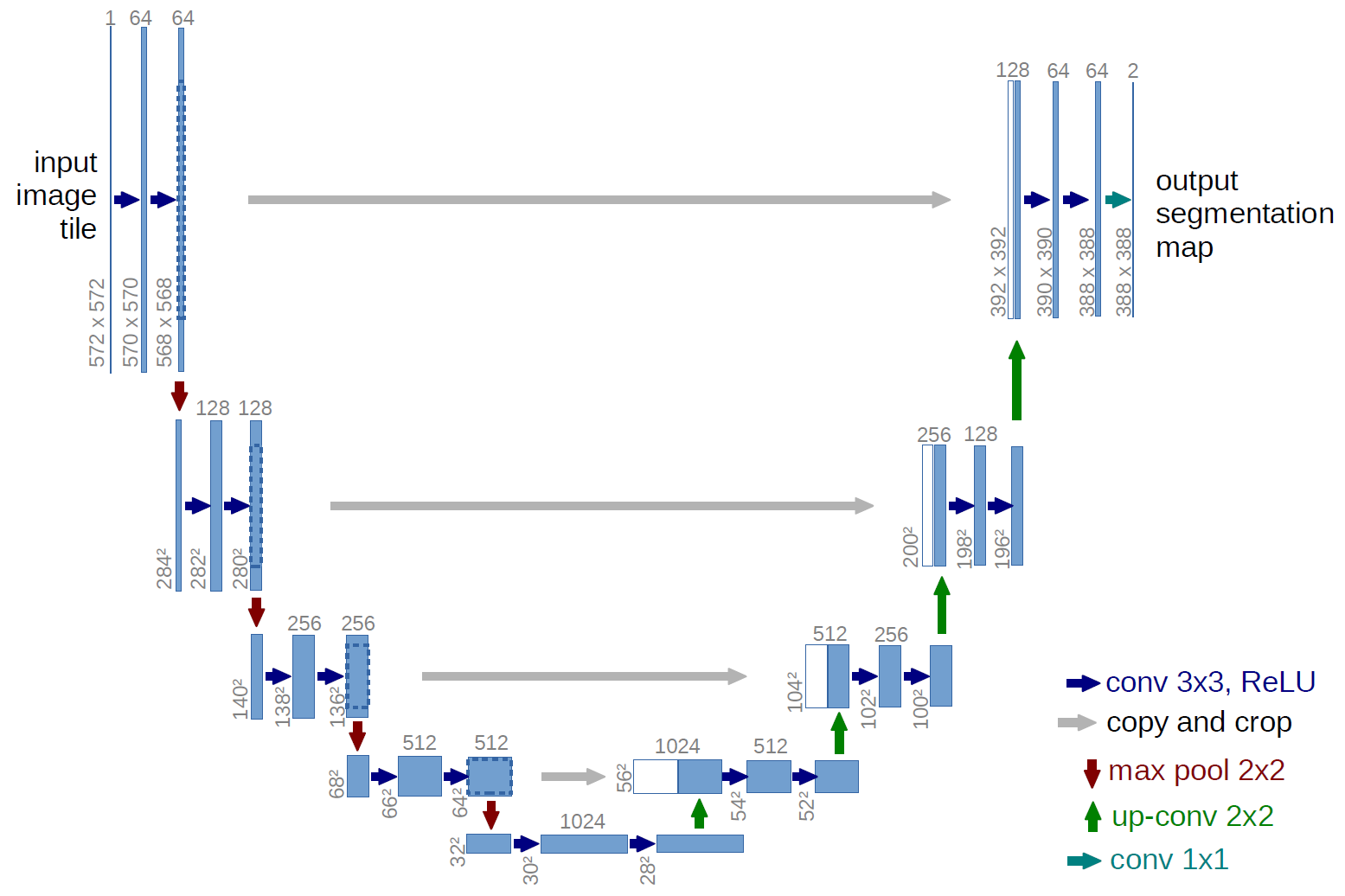}
\caption{A U-Net architecture}
\label{fig:myimage}
\end{figure}

It consists of four blocks, in each of which two convolutions are sequentially applied to extract increasingly detailed features, followed by a Max Pooling operation to halve the feature map size.
Then all the obtained feature maps are passed already to the decoder through the bottleneck - the deepest part of the network, where the maximum compression of information takes place before it will be restored in the decoder. 
The decoder consists of blocks similar to those of the encoder, but instead of Max Pooling a transposed convolution is used, which doubles the spatial size of the feature map. Through skip connection mechanisms, the features from the encoder are concatenated with the corresponding decoder level, which allows restoring spatial details lost during compression. 
The U-Net output generates a semantic mask corresponding to the original image dimensions. To reduce the computational complexity of the model, a lightweight U-Net is used, in which the number of extracted feature maps is reduced by a factor of 4. This modification allows to preserve the efficiency of segmentation while reducing the memory cost and training time.

\subsection{LinkNet}

LinkNet was introduced in 2017 as a semantic segmentation model designed for real-time image processing (Figure 2). The main idea is to efficiently transfer the extracted features between encoder and decoder layers using skip connections, which improves the accuracy and reduces the loss of spatial information.

\begin{figure}[H] 
\centering
\includegraphics[width=0.5\textwidth]{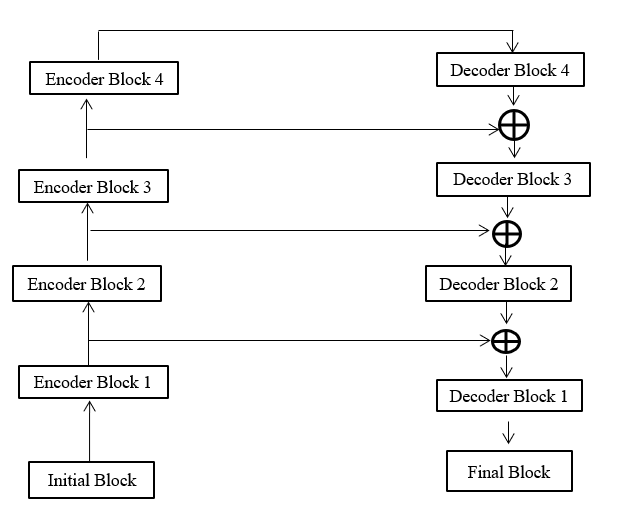}
\caption{A LinkNet architecture}
\label{fig:myimage}
\end{figure}

The LinkNet encoder starts with an initial block that applies convolution of the input image using a 7x7 kernel size and a 2x2 stride, and then performs max-pooling in a 3x3 area and a 2x2 stride. Next come 4 encoder blocks, each consisting of two residual blocks. With each level, the number of feature map channels is doubled, starting at 64. 
The decoder block, on the other hand, consists of two convolutions and one transpose convolution. With each decoder level, the number of feature map channels is also halved, starting at 512. The final block is responsible for bringing the feature map to the desired number of classes. With the transpose convolution of the 3x3 kernel reduces the feature map from 64 to 32, and then with the 2x2 kernel brings the size to the number of classes present in the dataset. 

\subsection{PSPNet}

Semantic Segmentation model Pyramid Scene Parcing Network (PSPNet) was submitted by Chinese University of Hong Kong and won the 2016 ImageNet Challenge. The first challenge of PSPNet was to improve Fully Convolutional network (FCN) model. With Pyramid Pooling Module (PPM), the neural network is able to capture both global and local features, which is especially special for complex images (Figure 3).

\begin{figure}[H] 
\centering
\includegraphics[width=0.5\textwidth]{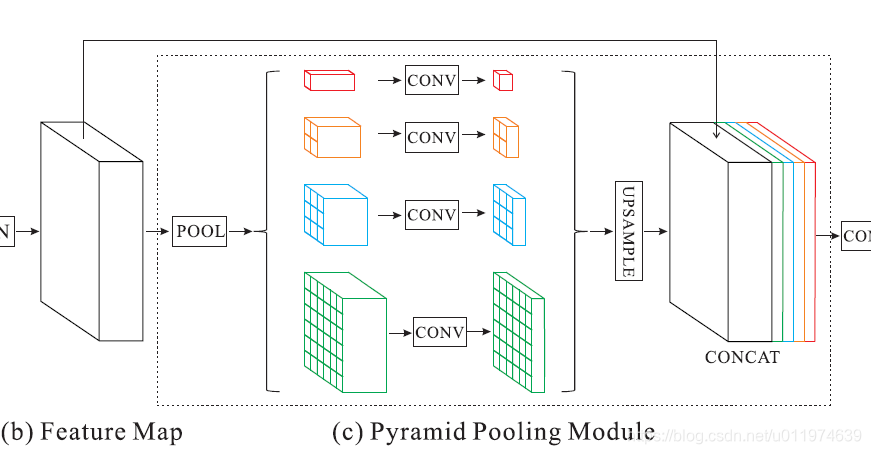}
\caption{A PSPNet architecture}
\label{fig:myimage}
\end{figure}

First, the input image is passed through the CNN where all possible features are extracted. Then from the final layer we pass them to the PPM. This module decomposes the feature maps into several layers with different resolutions, each of which extracts certain features. PPM creates a pyramid by performing Average pooling with different resolutions (1x1, 2x2, 3x3, 6x6). At the 1x1 level, the feature map is compressed into a single value for each channel to provide a global image context. At the 6x6 level, on the other hand, more localized features are extracted. 
After processing, the feature maps are combined using concatenation and the resulting maps are passed through a segmentation enhancement layer with a 3x3 convolution kernel.

\subsection{Metrics}

MeanIoU provides an objective measure of how well the model performs segmentation for each class. Unlike simple metrics such as accuracy, which can be uninformative in unbalanced data (e.g., when one class dominates), MeanIoU first calculates IoU (Intersection over Union) for each class separately and then averages these values. Here is the formula for calculating IoU:

\begin{equation}
\label{eq:iou}
\text{IoU} = \frac{|A \cap B|}{|A \cup B|}
\end{equation}

To calculate the IoU of one class, we need to determine the number of pixels simultaneously present in the predicted area A and true area B (Intersection), and find the union of the predicted and true areas. 
To improve the reliability of neural network accuracy data, F1 Score will be used in addition to MeanIoU. It has the advantage of being highly robust to class imbalance because F1-Score is the harmonic mean of Precision and Recall. 

\begin{equation}
\label{eq:prec_recall}
\text{Precision} = \frac{TP}{TP + FP} \quad
\text{Recall} = \frac{TP}{TP + FN}
\end{equation}

To calculate it, Precision and Recall must be found. Precision is a measure of how well it correctly classifies positive objects and Recall is a measure of how well the model finds all positive objects. For multiclassification, F1 is usually calculated for each class and then averaged (macro-averaging). 

\begin{equation}
\label{eq:f1}
F_1 = 2 \times \frac{\text{Precision} \times \text{Recall}}{\text{Precision} + \text{Recall}}
\end{equation}

Both metrics have been successfully applied in studies with semantic segmentation \cite{wu2022historicalmaps} \cite{zhou2018unetpp}. They are well suited for multi-class segmentation tasks, as they allow us to objectively evaluate the quality of recognizing each class, including rare classes. 

\maketitle
\section{Implementation and results}

\subsection{Preprocessing}

Before training the neural network, the images were prepared and converted into a format suitable for further training. To simplify the computation, we convert the images to grayscale using the OpenCV library and normalize the color range from [0, 255] to [0, 1]. Then, using the sklearn library, we sort the mask classes starting at 0. This is important for stable model performance.
The neural network function, which we have created, expects the input data tensor to be 4-dimensional (number of classes, width, height, number of channels). In the case of grayscale images, the value of the 3rd dimensional (if counting from 0), will be equal to 1. In the case of color images, it will be equal to 3. For this purpose we also use the sklearn library. The last step is to convert the input mask data arrays into one-hot-encoding format. It converts each pixel into a vector where the element corresponding to the true class of the pixel is set to 1, and the other elements take the value 0. This contributes to the efficient prediction of the neural network, furthermore, the Categorical Crossentropy loss function is exactly in this format and waits for the training data. 

\subsection{Training}

To properly compare the results, all six training sessions were run under identical conditions in the Google Colab environment. The software used included Keras 2.18.0, TensorFlow 2.18.0, and Python 3.11.11. An NVIDIA T4 graphics card (accessed via Colab Pro) and 50 GB of RAM were used to speed up the computations.

\begin{table}[H]
    \centering
    \renewcommand{\arraystretch}{1.2} 
    \begin{tabular}{ll}
        \toprule
        \textbf{Hyperparameter} & \textbf{Parameter Values} \\
        \midrule
        Input\_shape        & [128, 128] \& [192, 192] \\
        Optimizer           & Adam \\
        Loss                & categorical cross entropy \\
        Pretrained weights  & False \\
        Learning rate       & 0.001 \\
        Decoder (encoder, bottleneck, decoder) & 0.1, 0.3, 0.2 \\
        Batch Size          & 16 \\
        Epochs              & 100 \\
        Activation function & ReLU \\
        \bottomrule
    \end{tabular}
    \caption{MODEL HYPERPARAMETER VALUES}
    \label{tab:hyperparameters}
\end{table}

For PSPNet, ResNet18 was used as the CNN. The input image size was fixed at 128×128 for U-Net (light) and LinkNet and 192×192 for PSPNet. This difference arises because of the peculiarity of the PPM and the size of its Average pooling filters. All hyperparameters used are listed in Table 1. As mentioned before, the first dataset contains 4 classes (background, foundation, walls and roof) and consists of 2149 images, while the second dataset has 44 classes (one cube-one class) and consists of 870 images. Validation and test data constitute 10\% each of the total volume of each dataset. Of the 2149 images, 215 are validation images and another 194 are test images. In the second dataset, 88 images are for validation and 79 for testing. To reduce overtraining of the neural network in Unet(light) and LinkNet, we added Dropout. Its essence is that at each training step a certain number of neurons together with their connections are randomly switched off. This forces the network not to rely on specific neurons, but to distribute the information representation over a set of nodes.
To begin with, the results of the neural networks were examined based on the first dataset. All three models performed well with MeanIou accuracy above 60\% and F1 Score above 70\%. The best result was demonstrated by U-Net(light) with an accuracy of 0.7789 MeanIoU. The results of LinkNet and PSPNet were lower with 0.6118 and 0.6116, respectively. All the results of the metrics for the second dataset are presented in Table 2.

\begin{table}[H]
    \centering
    \renewcommand{\arraystretch}{1.2} 
    \begin{tabular}{lcccc}
        \toprule
        \textbf{Model} & \textbf{MeanIoU (I)} & \textbf{F1 Score (I)} & \textbf{MeanIoU (II)} & \textbf{F1 Score (II)} \\
        \midrule
        U-Net (light) & 0.7789 & 0.8720 & 0.1652 & 0.2508 \\
        LinkNet       & 0.6118 & 0.7478 & 0.0851 & 0.1955 \\
        PSPNet        & 0.6116 & 0.7480 & 0.1068 & 0.1665 \\
        \bottomrule
    \end{tabular}
    \caption{COMPARISON OF METRICS FOR ALL MODELS}
    \label{tab:twodatasets}
\end{table}

For the second dataset (with 44 classes), the results were much lower. The highest score was shown by the U-Net(light) model with MeanIoU 0.1652. LinkNet showed the worst result with MeanIoU 0.0751 and IoU for classes did not exceed 0.1580 This indicates problems with the distinction between cubes and poor preservation of the boundaries between them. At the same time, the background (class 0) always showed an accuracy above 80\%. The highest accuracy was observed for class 16, with an IoU of 0.357. However, for some classes, the IoU values reached 0.00.  The trend of training loss for each model is shown in Figure 4.

\begin{figure}[ht]
    \centering
    \begin{subfigure}[b]{0.3\textwidth}
        \centering
        \includegraphics[width=\textwidth]{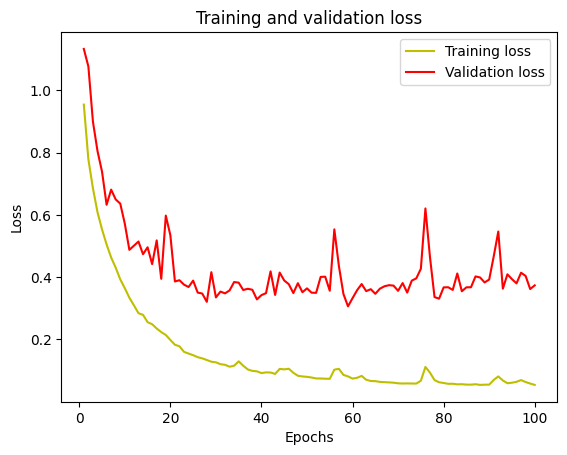}
        \caption{U-Net(I)}
        \label{fig:sub1}
    \end{subfigure}
    \hfill
    \begin{subfigure}[b]{0.3\textwidth}
        \centering
        \includegraphics[width=\textwidth]{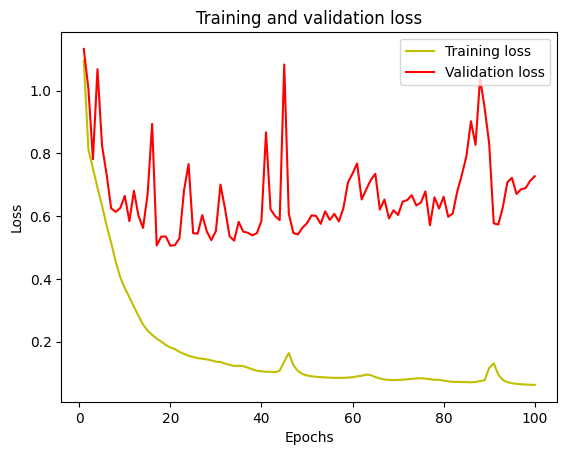}
        \caption{LinkNet(I)}
        \label{fig:sub2}
    \end{subfigure}
    \hfill
    \begin{subfigure}[b]{0.3\textwidth}
        \centering
        \includegraphics[width=\textwidth]{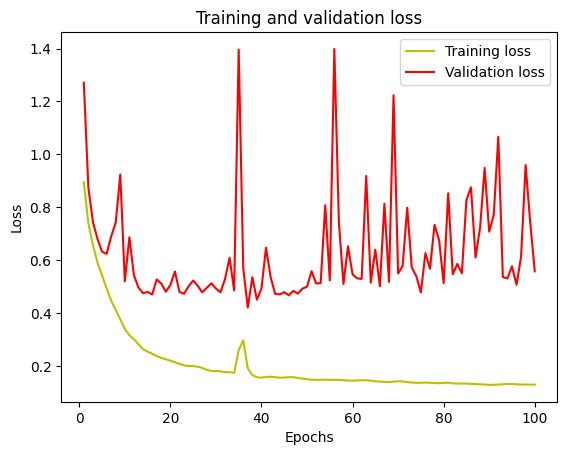}
        \caption{PSPNet(I)}
        \label{fig:sub3}
    \end{subfigure}

    \begin{subfigure}[b]{0.3\textwidth}
        \centering
        \includegraphics[width=\textwidth]{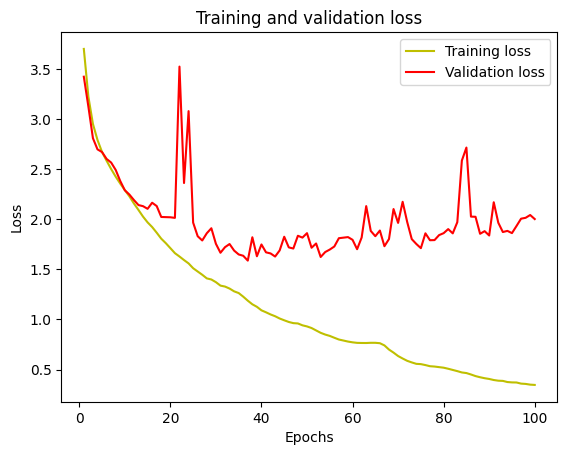}
        \caption{U-Net(II)}
        \label{fig:sub4}
    \end{subfigure}
    \hfill
    \begin{subfigure}[b]{0.3\textwidth}
        \centering
        \includegraphics[width=\textwidth]{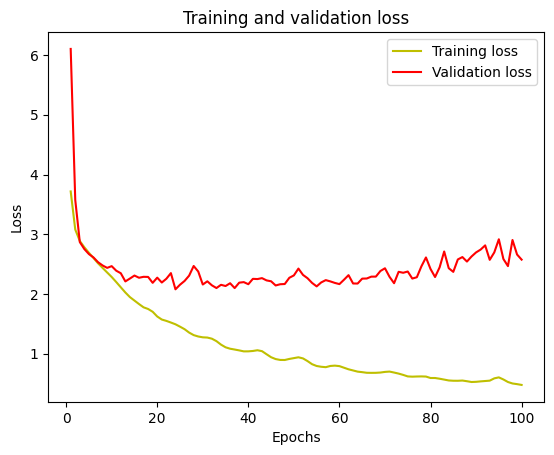}
        \caption{LinkNet(II)}
        \label{fig:sub5}
    \end{subfigure}
    \hfill
    \begin{subfigure}[b]{0.3\textwidth}
        \centering
        \includegraphics[width=\textwidth]{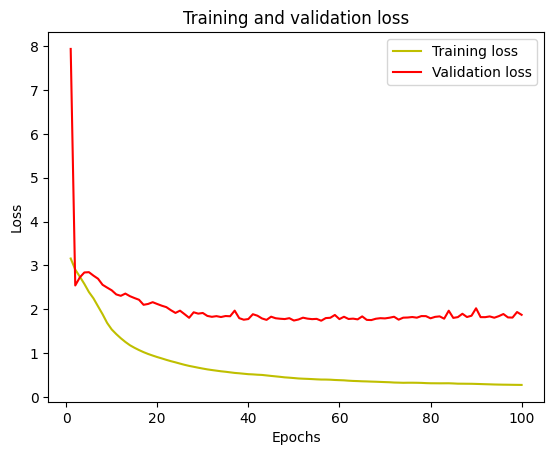}
        \caption{PSPNet(II)}
        \label{fig:sub6}
    \end{subfigure}

    \caption{The top three plots show the training and validation loss curves for the first dataset (4 classes), while the bottom three plots represent the second dataset (44 classes).}
    \label{fig:3x2images}
\end{figure}

The training time of the neural networks did not exceed 15 minutes for all models except for PSPNet, which took 73 minutes on the first dataset. For the second dataset, the execution environment was changed from a T4 graphics card to a more powerful one with L4 to increase memory, and the training time was 22 minutes due to the larger number of classes and the need for more computational resources.
The low performance and class imbalance on the second dataset was due to the small number of diverse images. The same reason affected the whole neural network training in general. In the plots of Figure 5, it can be seen that the accuracy on training reaches high values (above 0.8 and 0.5 MeanIoU respectively), but the accuracy on validation dataset decreased significantly, which is a clear sign of overfitting. This means that the models have memorized the training data well, but fail to cope with new images that they have not seen before. Solutions to overfitting problems have been proposed in \cite{ying2019overfitting} and are planned to be applied in the future. In Figure 6, you can see the predictions obtained from all models.

\begin{figure}[H]
    \centering
    \begin{subfigure}[b]{0.3\textwidth}
        \centering
        \includegraphics[width=\textwidth]{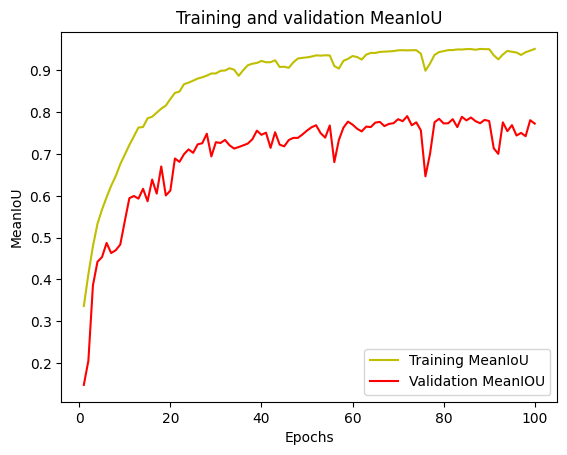}
        \caption{U-Net(I)}
        \label{fig:sub1}
    \end{subfigure}
    \hfill
    \begin{subfigure}[b]{0.3\textwidth}
        \centering
        \includegraphics[width=\textwidth]{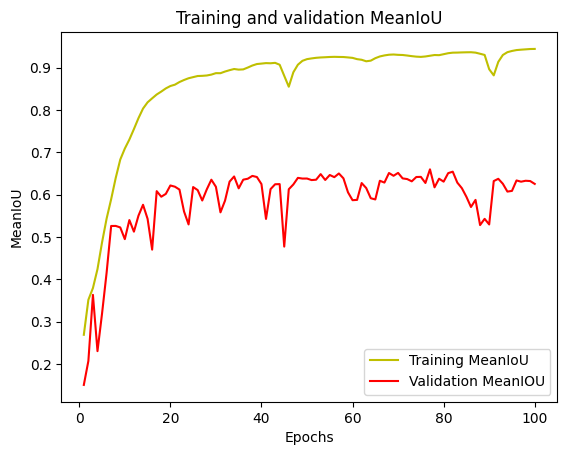}
        \caption{LinkNet(I)}
        \label{fig:sub2}
    \end{subfigure}
    \hfill
    \begin{subfigure}[b]{0.3\textwidth}
        \centering
        \includegraphics[width=\textwidth]{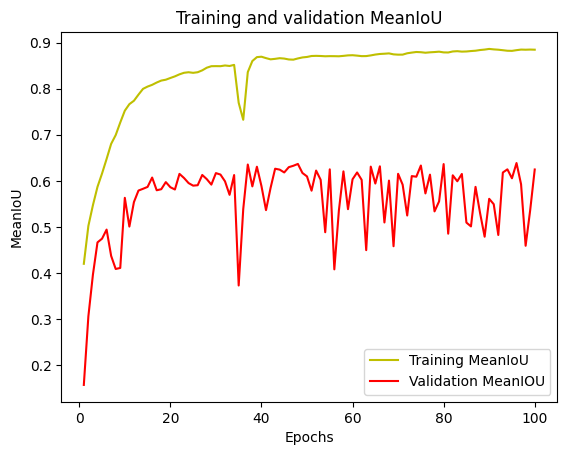}
        \caption{PSPNet(I)}
        \label{fig:sub3}
    \end{subfigure}

    \begin{subfigure}[b]{0.3\textwidth}
        \centering
        \includegraphics[width=\textwidth]{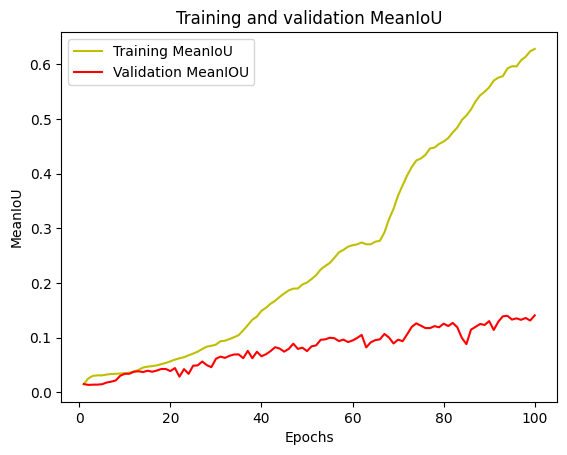}
        \caption{U-Net(II)}
        \label{fig:sub4}
    \end{subfigure}
    \hfill
    \begin{subfigure}[b]{0.3\textwidth}
        \centering
        \includegraphics[width=\textwidth]{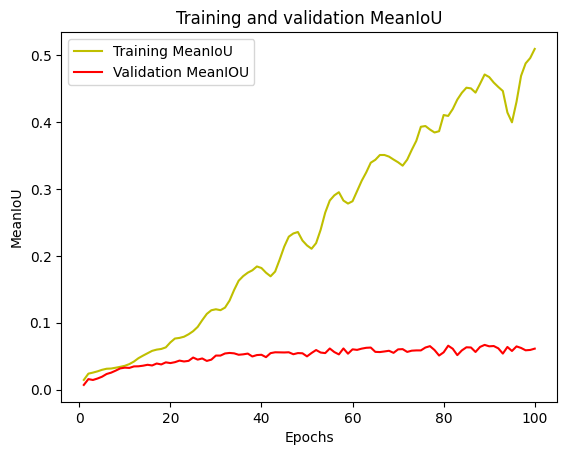}
        \caption{LinkNet(II)}
        \label{fig:sub5}
    \end{subfigure}
    \hfill
    \begin{subfigure}[b]{0.3\textwidth}
        \centering
        \includegraphics[width=\textwidth]{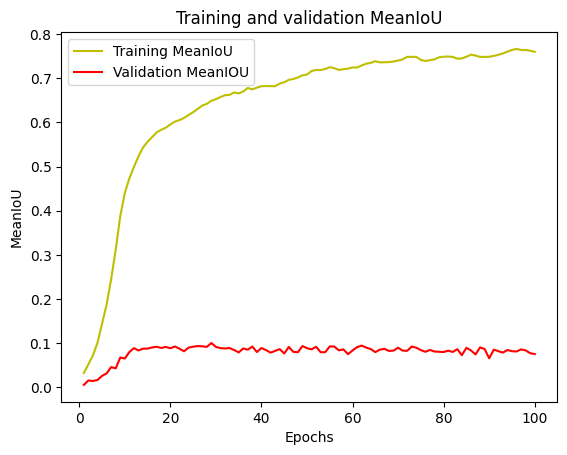}
        \caption{PSPNet(II)}
        \label{fig:sub6}
    \end{subfigure}

    \caption{The top three plots show the training and validation OneHotMeanIoU accuracy curves for the first dataset (4 classes), while the bottom three plots represent the second dataset (44 classes).}
    \label{fig:3x2images}
\end{figure}

\begin{figure}[H]
    \centering
    \begin{subfigure}[b]{0.3\textwidth}
        \centering
        \includegraphics[width=\textwidth]{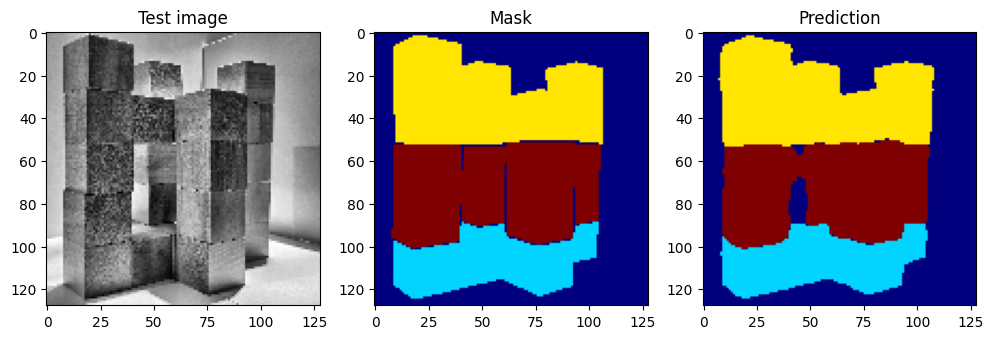}
        \caption{U-Net(I)}
        \label{fig:sub1}
    \end{subfigure}
    \hfill
    \begin{subfigure}[b]{0.3\textwidth}
        \centering
        \includegraphics[width=\textwidth]{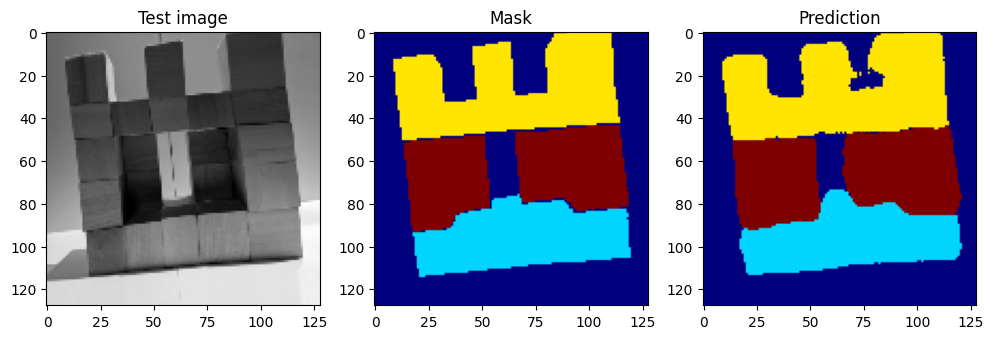}
        \caption{LinkNet(I)}
        \label{fig:sub2}
    \end{subfigure}
    \hfill
    \begin{subfigure}[b]{0.3\textwidth}
        \centering
        \includegraphics[width=\textwidth]{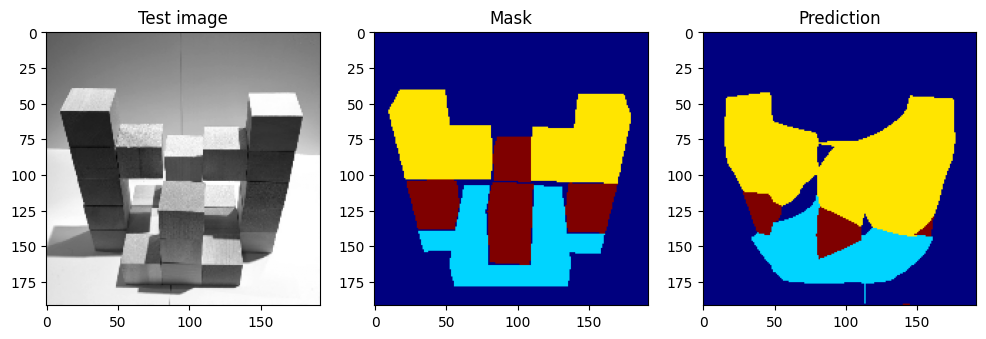}
        \caption{PSPNet(I)}
        \label{fig:sub3}
    \end{subfigure}

    \begin{subfigure}[b]{0.3\textwidth}
        \centering
        \includegraphics[width=\textwidth]{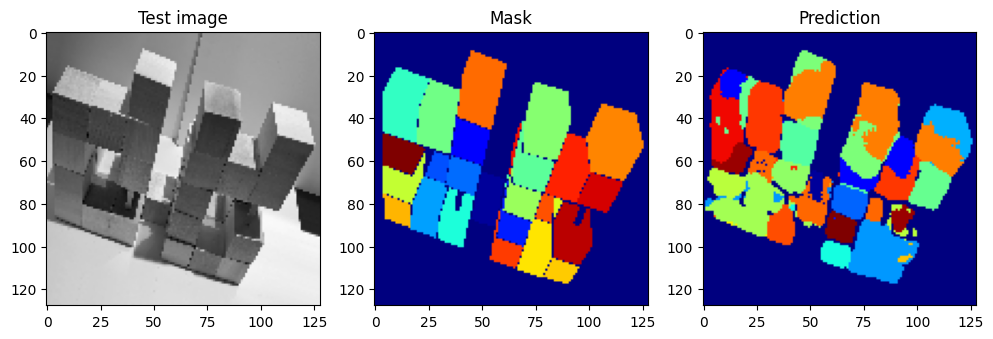}
        \caption{U-Net(II)}
        \label{fig:sub4}
    \end{subfigure}
    \hfill
    \begin{subfigure}[b]{0.3\textwidth}
        \centering
        \includegraphics[width=\textwidth]{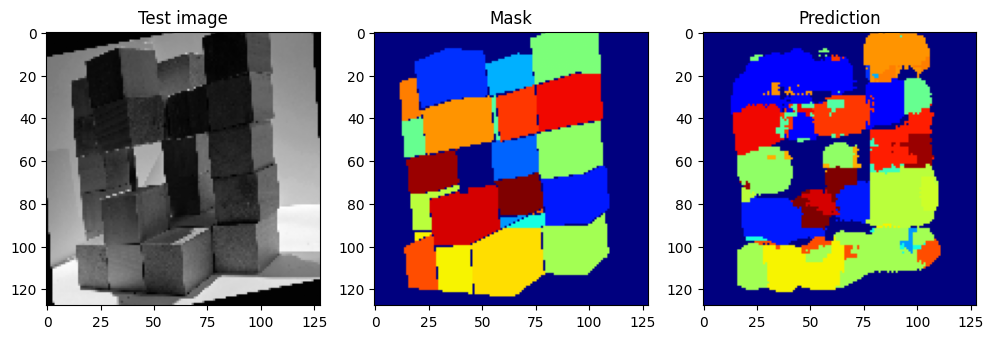}
        \caption{LinkNet(II)}
        \label{fig:sub5}
    \end{subfigure}
    \hfill
    \begin{subfigure}[b]{0.3\textwidth}
        \centering
        \includegraphics[width=\textwidth]{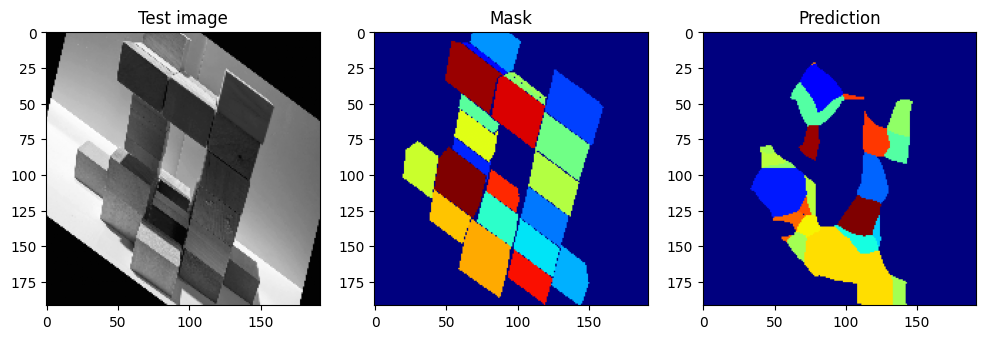}
        \caption{PSPNet(II)}
        \label{fig:sub6}
    \end{subfigure}

    \caption{Comparison of label mask with predictions from all models}
    \label{fig:3x2images}
\end{figure}

\maketitle
\section{Conclusion}
In this paper, a comparative analysis of three neural network architectures used for the semantic segmentation task is carried out on the basis of specially prepared datasets with images of houses built from wooden cubes.
The first dataset contained 4 classes (background, foundation, walls, roof) and was used for basic model evaluation and data quality checking. The second dataset consisted of 44 classes, where each individual cube was considered an independent class.
Three models, U-Net (light), LinkNet and PSPNet, using the same hyperparameters, were trained and tested. The accuracy was evaluated using MeanIoU and F1 Score metrics. According to the experimental results, U-Net (light) showed the best accuracy on both datasets: 78\% by MeanIoU and 87\% by F1 Score on the first dataset and 17\% and 25\%, respectively, on the second dataset.
These results confirm that U-Net (light), despite its reduced amount of input filters on convolution layers, retains the ability to segment building blocks efficiently. However, the low performance on the second dataset indicates the difficulty of the task of accurately selecting individual cubes due to the high similarity of classes and the limited amount of diverse data.
The results obtained can be the basis for further development of algorithms for automatic generation of a step-by-step construction plan and automated assembly of houses from cubes using manipulators. In the future, this technology can be adapted to create automatic design systems in real construction, which will optimize costs, reduce design time and improve the accuracy of erecting objects.
In future work, both datasets are planned to be significantly expanded: the first and second datasets will be increased to 3000 images each. To address the problem of overfitting and improve data diversity, an Early stopping system will also be implemented, which will stop training if the accuracy of the model stops growing. In addition, L1 and L2 regularization methods will be added to the neural network code along with the already used Dropout. To improve class difference, each class will receive precise markings on the surface of the cubes. The next key step in the project is to create algorithms to generate a step-by-step plan for building houses from cubes. This may require collaboration with other researchers and experts to ensure the practicality and accuracy of the generated plans.

\end{document}